\documentclass{article}

\usepackage{PRIMEarxiv}

\usepackage[utf8]{inputenc} 
\usepackage[T1]{fontenc}    
\usepackage{hyperref}       
\usepackage{url}            
\usepackage{booktabs}       
\usepackage{amsfonts}       
\usepackage{nicefrac}       
\usepackage{microtype}      
\usepackage{lipsum}
\usepackage{fancyhdr}       
\usepackage{graphicx}       
\graphicspath{{media/}}     

\title{TransformLLM: Adapting Large Language Models via LLM-Transformed Reading Comprehension Text}

\author{
  Iftach Arbel\\
\textit{School of Mathematical Sciences} \\
\textit{Tel Aviv University}\\
Tel Aviv, Israel \\
  \texttt{i.arbel84@gmail.com} \\
   \And
  Yehonathan Refael \\
\textit{Department of Electrical Engineering} \\
\textit{Tel Aviv University}\\
Tel Aviv, Israel \\
refaelkalim@mail.tau.ac.il\\
\And
  Ofir Lindenbaum \\
  \textit{The Faculty of Engineering} \\
\textit{Bar Ilan University}\\
Ramat Gan, Israel \\
ofir.lindenbaum@biu.ac.il\\
}
\usepackage{times}
\usepackage{latexsym}

\usepackage[utf8]{inputenc}

\usepackage{microtype}

\usepackage{inconsolata}

\usepackage{graphicx}
\usepackage{comment}

\usepackage{multirow} 
\usepackage{array}
\usepackage{tabularx}
\usepackage{makecell}
\usepackage[numbers]{natbib} 
\usepackage{cleveref}
\usepackage{float}
\usepackage{placeins}

\usepackage{booktabs}
\usepackage{adjustbox}

\newcommand{\bbo}{{\bf O}}

\newcommand{\bbv}{{\bf V}}

\newcommand{\bbk}{{\bf K}}
\newcommand{\bbq}{{\bf Q}}

\begin{document}
\maketitle

\begin{abstract}

Large Language Models (LLMs) have shown promise in highly-specialized domains, however challenges are still present in aspects of accuracy and costs. These limitations restrict the usage of existing models in domain-specific tasks. While fine-tuning pre-trained models have shown promising results, this process can be computationally expensive and require massive datasets of the specialized application in hand. In this work, we bridge that gap. We have developed Phi-2-Legal and Mistral-Legal-7B, which are language models specifically designed for legal applications. These models are based on Phi-2 and Mistral-7B-v0.1, and have gone through continued pre-training with over 500 million tokens of legal texts. Our innovative approach significantly improves capabilities in legal tasks by using Large Language Models (LLMs) to convert raw training data into reading comprehension text. Our legal LLMs have demonstrated superior performance in legal benchmarks, even outperforming models trained on much larger datasets with more resources. This work emphasizes the effectiveness of continued pre-training on domain-specific texts, while using affordable LLMs for data conversion, which gives these models domain expertise while retaining general language understanding capabilities. While this work uses the legal domain as a test case, our method can be scaled and applied to any pre-training dataset, resulting in significant improvements across different tasks. These findings underscore the potential of domain-adaptive pre-training and reading comprehension for the development of highly effective domain-specific language models.
\end{abstract}

\section{Introduction}
Large Language Models (LLM) domain-adaptive pre-training, also known as continued pre-training on domain-specific corpora \citep{gururangan2020don}, is a technique that has been proven effective in adapting large language models (LLMs) to specific domains \citep{yao2021adapt,cheng2022snapshot}. This approach allows LLMs to leverage their general language understanding capabilities while incorporating domain-specific knowledge, which can benefit downstream domain-specific tasks at reduced costs \citep{liu2019roberta,rosenbaum2022clasp, rosenbaum2022linguist}.

In this process, the LLM is further pre-trained using raw data from the specific domain, such as biomedicine, finance, or law. This helps the LLM gain domain knowledge, which is demonstrated by its improved performance in fine-tuning and knowledge probing evaluations within those domains \citep{lee2020biobert,beltagy2019scibert,chalkidis2020legal}. However, a notable drawback is that continued pre-training on raw domain corpora can lead to a significant drop in the LLM's prompting performance, potentially due to the specialized nature of the domain-specific data \citep{gu2021domain}. Despite this trade-off, domain-adaptive pre-training remains a promising approach for adapting LLMs to specific domains, capitalizing on their general language understanding capabilities while tailoring them to domain-specific tasks and knowledge. Ongoing research efforts aim to mitigate the potential negative impacts on prompting performance while maximizing the benefits of domain-specific knowledge acquisition \citep{golchin2023mask,rozner2024knowledge}.

The notion of reading comprehension was suggested in \citep{cheng2023adapting}, where instead of continuing to train a large language model on domain-specific raw data, the raw texts be converted into reading comprehension materials. In this approach, each text is followed by related tasks, transitioning the model from a "reading" phase to a "comprehension" phase. These tasks, in a question-answer format, enhance the model's ability to respond to questions by simulating human learning practices. 

We introduce novel methods to expose the models to a corpus during training, blending a variety of legal reading comprehension tasks, as well as general language data. To demonstrate the performance of our method, we utilize Phi-2 and Mistral-7B as base models, which were further pre-trained on 500 million tokens of legal corpus. Our new legal LLMs present state-of-the-art performance on legal benchmarks, suppressing models trained on larger corpora with significantly more resources.

Our main contributions are: (i) Utilizing LLMs, to transform raw text to reading comprehension text that is used for continued pre-training of LLMs in legal domain tasks. (ii) Develop an extended evaluation scheme for legal LLMs. Existing legal benchmarks are currently fragmented and constructed for classification models with multiple question responses. Our evaluation protocol adapts MMLU \citep{hendrycks2020measuring} (legal subsets) and LexGLUE \citep{chalkidis2021lexglue} for use with generative, GPT-style \citep{radford2018improving} transformer \citep{vaswani2017attention} models. While the focus of this work is on the legal domain, both the transformation and evaluation protocols are easily applicable to other domains, including finance, biology, and more.

\section{Using LLMs to Transform Raw Text}

Building upon the foundation of AdaptLLM \citep{cheng2023adapting}, which converts raw legal text into reading comprehension tasks, we draw from the concept of human learning through reading comprehension. This approach, where practice after reading improves the ability to answer questions based on acquired knowledge, inspired our work. Rather than continuing to train large language models on raw domain-specific corpora, AdaptLLM proposes converting the raw text into structured reading comprehension tasks, with each passage followed by questions.

While AdaptLLM leverages a set of rules and heuristics to perform this transformation, its reliance on such methods poses limitations, especially in the resulting data qualityquality of the resulting data. These challenges highlight a critical need for more sophisticated text transformation techniques \citep{liu2024evolutionheuristicsefficientautomatic,sharma2024textqualitybasedpruningefficient}. Our solution addresses this by leveraging large language models (LLMs) to generate high-quality training data. With the decreasing costs of LLM inference, we can move beyond structured heuristics, using LLMs to efficiently create comprehensive reading comprehension datasetscreate comprehensive reading comprehension datasets efficiently.

To improve text quality, we designed a prompt database that guides the model’s capabilities. LLMs were tasked with generating answers and additional questions, additional questions, and transforming the raw legal texts based on tailored prompts. Through further refinement and post-processing, we developed a superior legal reading comprehension dataset, offering enhanced performance for domain adaptation.

We primarily used open-source models ranging from 7B to 70B for data transformation. These models were selected based on factors like cost and operational efficiency. Upon reviewing the outputs of these open-source models comparedin comparison to more advanced proprietary models like those from OpenAI and proprietary Mistral models, we observed no significant differences in quality for our transformation task. However, to ensure a diverse data distribution and to benefit from knowledge distillation of the most powerful models, we also transformed a portion of the data using state-of-the-art proprietary (closed-source) models.

Some transformations were also applied on the general, non-legal data to generate Chain-of-Thought (CoT) data and improve the reasoning capabilities of the model, which we find crucial in the legal domain. For the same reason, we incorporated math and coding data in the training set, striving to boost logical and inference capabilities.

\section{Data Collection and Processing}

Our data collection focused primarily on English-language legal texts, drawing heavily from the United States, which follows the common law system. We also included materials from Canada and the United Kingdom, both of which also adhere to common law principles. This emphasis on jurisdictions with common law traditions ensures that our dataset aligns closely with the legal usage specific to the United States, which is the primary focus of our model. Through meticulous curation and rigorous cleaning procedures, we compiled a comprehensive corpus tailored to capture the intricacies of legal language within the United-States Federal-jurisdiction.

Throughout the development of the model and the data collection process, our goal was not to expose the model to all existing legal data. Instead, we focused on providing the model with a strong foundation of legal knowledge, background, understanding, and reasoning abilities. Our aim is for the model to be able to handle various legal tasks, including document drafting, reviews, and answering questions, by equipping it with these tools. However, if you ask the model about specific data such as cases or laws, it may provide inaccurate information. In such cases, Retrieval-Augmented Generation (RAG) is the recommended solution. Utilizing a robust legal LLM, along with a retrieval model and a comprehensive database, will yield reliable and solid results.

The main sources of legal data was raw text from the FreeLaw subset of The Pile \citep{gao2020pile} and Pile of Law \citep{henderson2022pile}. The Pile dataset does not have any indexing, therefore we simply sample data from it, while using word count to evaluate the number of raw tokens we attained. Pile of Law, on the other hand, does index the data by instances, so we could sample data that we find appropriate, including contracts, SEC filing, and legal memos to name a few. This indexing also allowed to avoid certain data instances, such as Congressional hearings and European laws.

In order to avoid regression of general language capabilities during the fine-tuning process, we integrated data from the original training distribution, a strategy supported by previous studies \citep{wu2024continual, cheng2023adapting}. We introduced widely available "general" instruction data from various sources, including chain-of-thought (CoT), chat, code, and general instruction datasets. The datasets were sampled from a diverse range of resources, ensuring a broad spectrum of language usage and contexts, thereby preserving the model's general language capabilities while enhancing its performance in the legal domain. 

The set of datasets used in this paper is following presented in Table \ref{data-sources}.

\begin{table}[h!]
  \centering
  \begin{tabular}{llll}
    \hline
    \textbf{Domain}   & \textbf{Dataset}    & \textbf{Tokens} & \textbf{License} \\
    \hline
    Legal             & The Pile (FreeLaw)  &  300M           & MIT  \\
    Legal             & Pile of Law         &  180M           & CC-BY-NC-SA-4.0  \\
    Legal             & USClassActions      &  20M            & GPL-3.0  \\
    Math (CoT)        & AQUA-RAT            &  5M             & Apache-2.0  \\
    Commonsense (CoT) & ECQA                &  4M             & Apache-2.0  \\
    Reasoning (CoT)   & EntailmentBank      &  3M             & Apache-2.0  \\
    Chat              & UltraChat           &  140M           & MIT  \\
    Code              & Code-Feedback       &  60M            & Apache-2.0 \\
    Instruction       & OpenOrca            &  300M           & MIT  \\
    \hline
  \end{tabular}
  \centering
  \caption{\label{data-sources}
    A list of used data sources.
  }
\end{table}
\FloatBarrier

Examples from the training data are shown in the Table \ref{tab:explanations}, in Training Samples Example section \ref{app::TrainingSamplesExample}, in the appendix.


\begin{table*}
  \centering
  \begin{tabular}{l c c c c c c}
    \hline
     & \multicolumn{3}{c}{\textbf{MMLU}} & \multicolumn{3}{c}{\textbf{LexGLUE}} \\
    \cmidrule(r){2-4}  \cmidrule(r){5-7}
     & \makecell{International \\ Law} & \makecell{Juris-\\prudence} & \makecell{Professional \\ Law} & LEDGAR & \makecell{Case\\HOLD} & \makecell{Unfair \\ ToS} \\
    \hline
    \multicolumn{7}{c}{\textbf{3B Models}} \\
    \hline
         Phi-2 & 0.661 & 0.620 & 0.379 & 0.143 & 0.310 & 0.233 \\
         \textbf{Phi-2-Legal} & \textbf{0.667} & \textbf{0.711} & \textbf{0.417} & \textbf{0.603} & \textbf{0.580} & \textbf{0.385} \\
        \hline
    \multicolumn{7}{c}{\textbf{7B Models}} \\
    \hline
         Mistral-7B & 0.736 & 0.694 & 0.412 & 0.506 & 0.563 & 0.366 \\
         AdaptLLM \ & 0.570 & 0.528 & 0.361 & 0.463 & 0.500 & 0.513 \\
         Saul-7B & 0.694 & 0.630 & \textbf{0.432} & 0.559 & 0.658 & 0.803 \\
         \textbf{Mistral-Legal-7B} & \textbf{0.811} & \textbf{0.712} & 0.427 & \textbf{0.739} & \textbf{0.778} & \textbf{0.806} \\
        \hline
  \end{tabular}
  \caption{\label{3b_7b_model_eval}
    Benchmark results for 3B and 7B Models}
\end{table*}

\section{Model Architecture and Training}

We have trained two versions of the legal model: Phi-2-Legal and Mistral-Legal-7B. As suggested by their names, these models are based on Phi-2 \citep{javaheripi2023phi} and Mistral-7B \citep{jiang2023mistral}. We selected these models because they demonstrate cutting-edge performance, are available for commercial use, and are well-supported by inference libraries (vLLM  \citep{kwon2023efficient}, etc.) and for server-less deployment (Fireworks, Together, etc.). 


\subsection{Training considerations} 
To save on resources and consider the very limited availability of GPUs, we opt to train the models using LoRA \citep{hu2021lora}, avoiding full parameter update. Lora is a parameter-efficient fine-tuning (PEFT) technique that has beenParameter-Efficient Fine-Tuning (PEFT) technique, proven to match the results of full-parameter updates while requiring significantly fewer training resources (Note that any state-of-the-art variants of LoRA \cite{refael2024adarankgradadaptivegradientrankmoments,Chen2023,Wang2023,Xu2023} may be used as an alternative). Considering the vast training data and project scope, we train a considerable amount of parameters. Both models used a LoRA $r$ = 64, and updated all attention components ($\bbq$, $\bbk$, $\bbv$, $\bbo$), as well as the feed-forward (FF) layers. These models can support context lengths of up to 2K and 4K for Phi-2 and Mistral-7B, respectively\footnote{Mistral context length is without "sliding-window attention".}. 

In order to improve training efficiency, a common technique involves packing the training data. Packing refers to concatenating examples in a greedy fashion until the full context length is reached, while using an EOS (end of sequence) token to separate them. We note that it is not possible to perfectly fit examples into the context length, so any overflow from one example is moved to the next training example. While this technique generally works well for model pre-training and most fine-tuning scenarios, it is unsuitable for our case. Since we are focused on reading comprehension, where the model is presented with raw text followed by a series of questions, cutting examples poses a risk to the capabilities of the fine-tuned model. Therefore, we use a packing mechanism which concatenates examples without cutting any of them. Achieving a perfect concatenation is not possible, as this problem is essentially the bin-packing problem \citep{martello1990lower, korte2018bin}, which is NP-hard. However, a good approximation is simply sorting the data by length and packing examples using a greedy algorithm. Using this algorithm, we compressed the training set by 70\verb|%|-80\verb|%|, depending on context length.

\section{Evaluation}
We evaluate the models on MMLU (legal subsets) and LexGLUE datasets. We aim for a simple and accessible evaluation scheme that is easy to understand and measures model accuracy. MMLU is typically evaluated using the Log probabilities of tokens, as in \citep{eval-harness}. However, this type of model evaluation has two main drawbacks: (1) Attaining raw requires setting up servers with GPUs and server-less inference providers,, and server-less inference providers as these are limited in the number of Log probabilities they output. (2) Measuring against Log probabilities may encounter issues due to tokenization mismatches. LexGLUE typicallynormally evaluates classification models rather than generative ones. Therefore, we adapt benchmark prompts for instruct-type models, detailing the various options and asking for the most suitable option to be selected. This means that models may be evaluated quickly and affordably using inference frameworks such as vLLM, or server-less inference providers. We also utilize recent advancements in decoding techniques, allowingadvancements in decoding techniques, which  us to define a closed listallows to define a closed-list of possible options. The result is a transparent and simple evaluation scheme suitable for , suitable to be used with chat-aligned models.

MMLU is a straightforward multiple-question benchmark. LexGLUE, on the other hand, has subsets that are simple multiple-question, while others have 8-100 label options. In LexGLUE, we only use the subsets that are suitable for use with generative models. For that, the EUR-LEX subset was not used as it only has numerical labels, not verbal, meaningful ones, while the SCOTUS subset was avoided as many of its instances are longer than a 4K token window; therefore, it, therefore is has very few usable data instances. Lastly, we did not use the ECtHR subsets, as they refer to proceedings brought to the European Court of Human Rights (ECtHR), and therefore rely on the European Convention on Human Rights,relies on European Convention on Human Rights which is a codified document more typical of civil law systems \citep{ECtHR2000}.

Our legal models were benchmarked compared to their underlying base models, Phi-2 and Mistral-7B, to measure the improvement achieved by continued pre-training. The Mistral-7B is also compared to the legal variant of AdaptLLM mode, which also uses continued pre-training using reading comprehension text. Additionally, we compare it to Saul-7B \citep{colombo2024saullm}, another recent legal model that uses at least x30 more training data and full-parameter update (compared to our LoRA training). We are unawarenot aware of legal models smaller than 7B parameters; therefore,, therefore the Phi-2 models are the only ones in this category. These benchmark results are presented in Table \ref{3b_7b_model_eval}.

Both classes of models show considerable improvement over their base models. Mistral-Legal-7B performs better in all subsets thancompared to AdaptLLM, highlightingwhich highlights the benefit of transforming raw data using LLMs, compared to the heuristic and regex rules. It also performs better than Saul-7B in five out of six subsets. We observed the most significant performance gains in the LexGLUE subsets. We suspect this is because LexGLUE is a more niche benchmark, receiving less attention from model developers. In contrast, the MMLU benchmark is highly popular, and the original models were already extensively optimized for it, making further improvements more challenging. Nevertheless, our method still enhancedmanaged to enhance results, with Phi-2-Legal outperforming the original Mistral-7B in all but one of the benchmark subsets.

\section{Conclusion}
In this work, we presented a framework for domain-specific adaptation of LLMs using continued pre-training. By training models in the legal domain, we have shown that it is possible to obtain high-performing models with relatively low resources. To the best of our knowledge, this is the first time this technique has been used. 

Future research could employ Reinforcement Learning from Human Feedback (RLHF) to enhance the model's alignment with human preferences. This would lead to improved generation capabilities and more refined outputs, advancing the applicability and efficacy of the model in diverse applications.

\section*{Limitations}
The models were evaluated using multiple-question benchmarks, which serve as proxies for their legal capabilities. However, a dedicated framework for evaluating its text generation capabilities, particularly in specific applications such as contracts and reviews, is necessary to obtain a comprehensive assessment. The models are not intended or able to provide factual information, they may generate information that is false or misleading, and reflect social and cultural biases from their training data, both the original pre-training data as well as our continued pre-training data.

\bibliographystyle{plainnat}
\bibliography{custom}
\clearpage

\appendix



\section{Ethics statements}
As with all large language models, there is an inherent risk of the model producing biased or toxic responses, which remains a significant concern and requires ongoing attention. Given its application in the legal domain, the language model's biases could have more severe implications than those in general usage scenarios. Moreover, mistakes and hallucinations produced by the model can result in costly errors to the end user.

\section{Training Samples Example}\label{app::TrainingSamplesExample}
In Table \ref{tab:explanations}, we provide examples of our text transformations, including prompts and responses by the LLM.

\begin{table*}[!htb]
\centering
\begin{tabular}{p{3cm}p{4.5cm}p{6.9cm}}
   
\hline
\textbf{Task Type} & \textbf{Prompt} & \textbf{Response} \\
\hline
Commonsense Reasoning & What is an explanation to this paragraph from the text, starting with: "We review the trial court's denial of a bond-reduction request under an abuse of discretion standard..."? & This paragraph explains the standard of review used by the court when assessing the trial court's decision to deny a bond-reduction request, which is an abuse of discretion standard, and clarifies that the trial court's decision must be arbitrary or unreasonable to be considered an abuse of discretion. \\
\hline
Word-to-Text & Provide a definition to these two legal terms from the text: "habeas corpus", "abuse of discretion". & "habeas corpus" - a writ requiring a person under arrest to be brought before a judge or into court, especially to secure the person's release unless lawful grounds are shown for their detention. "abuse of discretion" - a legal standard used to review a decision made by a lower court or administrative body, which occurs when the decision is arbitrary, capricious, or unreasonable. \\
\hline
Natural Language Inference & Does the sentence "The chapter 13 case was converted to a case under chapter 7 in July of 1993" entail the sentence "The record does not show the present value of the vehicle, but it is worth more than the \$300 remaining due under the chapter 13, and much less than the \$6,000 remaining due under the original contract"? & The two sentences are neutral with regard to each other. The first sentence discusses the conversion of the case from Chapter 13 to Chapter 7, while the second sentence talks about the value of the vehicle in question. \\
\hline
Summarization & Write a summary for this paragraph from the text, starting with: "District courts have jurisdiction to hear appeals from final judgments, orders, and decrees issued by bankruptcy judges...". & This paragraph explains the jurisdiction of district courts in hearing appeals from bankruptcy courts and the standards of review for legal conclusions and findings of fact. \\
\hline
\end{tabular}

\caption{Examples of raw text transformed to reading comprehension tasks, using LLMs \label{tab:explanations}.}
\end{table*}

\end{document}